\documentclass[a4paper,conference]{IEEEtran}
\usepackage{times}
\usepackage{epsfig}
\usepackage{graphicx}
\usepackage{amsmath}
\usepackage{amssymb}
\usepackage{float}
\usepackage{multirow}
\usepackage[pagebackref=true,breaklinks=true,letterpaper=true,colorlinks,bookmarks=false]{hyperref}

\ifCLASSINFOpdf
\else
\fi

\begin{document}

\title{Alcohol Consumption Detection from Periocular NIR Images Using Capsule Network}

\author{

\IEEEauthorblockN{Juan Tapia, Enrique Lopez Droguett and Christoph Busch}
\IEEEauthorblockA{da/sec-Biometrics and Internet Security Research Group, Hochschule Darmstadt, Germany\\
Department of Civil and Environmental Engineering, and Garrick Institute for the Risk Sciences,\\ University of California, USA.\\
Email: juan.tapia-farias@h-da.de, eald@g.ucla.edu, christoph.busch@h-da.de}
***This work has been presented at the ICPR 2022 conference for publication. \\Copyright may be transferred without notice, after which this version may no longer be accessible***.}

\maketitle

\begin{abstract}
This research proposes a method to detect alcohol consumption from Near-Infra-Red (NIR) periocular eye images. The study focuses on determining the effect of external factors such as alcohol on the Central Nervous System (CNS). The goal is to analyse how this impacts on iris and pupil movements and if it is possible to capture these changes with a standard iris NIR camera. This paper proposes a novel Fused Capsule Network (F-CapsNet) to classify iris NIR images taken under alcohol consumption subjects. The results show the F-CapsNet algorithm can detect alcohol consumption in iris NIR images with an accuracy of 92.3$\%$ using half of the parameters as the standard Capsule Network algorithm. This work is a step forward in developing an automatic system to estimate "Fitness for Duty" and prevent accidents due to alcohol consumption. 
\end{abstract}

\IEEEpeerreviewmaketitle

\section{Introduction}
Periocular NIR images have been mainly used to recognize subjects in controlled environments, and for soft-biometrics applications such as gender classifications, \cite{Periocular_tapia}.
Improvements and reductions in the cost of iris capture devices will witness broader applications and may be confronted with newer challenges. One unique challenge is identifying if a subject is under the effects of alcohol, drugs or sleep deprivation.
This area is known as Fitness for Duty (FFD) \cite{FFD,murphy} and allows to determine whether the person is physically or psychologically able to perform their task. \cite{murphy}.  

An Unfit subject is one under the influence of alcohol/drugs/sleepiness. Working under such conditions can affect workers' performance and increase the risk of accidents.
The impact of certain drugs (including alcohol) on the oculomotor system has been extensively studied in the literature \cite{BrownAdamsHaegerstrom-PortnoyEtAl1977, AroraVatsaSinghEtAl2012, Tomeo-ReyesRossChandran2016, M.C.RowbothamJones1987, M.C.RowbothamBenowitz.1984}. It is common knowledge that alcohol alone or mixed with cocaine or marijuana and sleep disruption are psychoactive agents that change the brain's functions. Such changes may be manifested in perception, memory, decision-making, attention, motor activity, and many other parts associated with the brain.
Alcohol consumption can lead to increased sleepiness and reduced alertness, even after the alcohol is no longer detectable in blood. Alcohol intoxication significantly impairs physical performance and cognitive functions.
Numerous studies in the literature have studied at the impact of these agents on basic skills  \cite{GennaroFerraraUrbaniEtAl2000}.

According to the World Health Organisation, there are  3.3 million deaths worldwide every year due to the harmful use of alcohol. This represent 5.9 \% of all deaths per year \footnote{\url{http://www.who.int/mediacentre/factsheets/fs349/en/}}. 

Duty of Care legislation exists in some countries such as Australia, the USA, Chile, and others that place responsibility on both employers and employees for maintaining a safe workplace. The employer's burden is to provide and maintain a safe and healthy workplace. The employee's responsibility is to undertake lifestyle management that ensures fitness for duty. As such, detecting people under the influence of alcohol is critical to prevent injures and accidents. Therefore, it is essential to have systems that can identify and robustly and efficiently detect the influence of alcohol in workers to ensure ”Fitness for Duty”. This kind of observation and analysis system are needed broadly with a high impact in several industries such as mining, health, logistics (driver), insurances, and others \cite{miner}.

Several commercial products are currently available to measure the FFD. For instances:

The Optalert\footnote{\url{https://www.optalert.com/}}, is an infrared reflectance (IR) oculography based on the principle that while an individual is tired, the central nervous system inhibits the muscle groups controlling eye and eyelid movements.

The Sobereye\footnote{\url{https://www.sober-eye.com/}}, is a portable device used to predict impairment caused by substance abuse or fatigue. It uses a smartphone attached to an opaque enclosure that fits a user's eyes to measure the Pupillary Light Reflex (PLR).

The PMI-FIT 2000\footnote{\url{https://www.pmifit.com/}}, uses eye-tracking and pupillometry to identify impaired physiological states due to fatigue and other factors, such as alcohol or drug use.

Some critical issues have been shown related to the type of device response and the possibility of impersonation. Most of these are portable devices (wearables) or using personal identification numbers. The impersonation is easy because they do not have a biometrical identification stage. They are mainly designed to detect risky events related to tiredness and correlation with alcohol, drugs and others.

This work focuses on detecting the presence of alcohol in a pro-activate fashion using iris information. It is essential to highlight that our proposal will not perform a traditional analysis based on the measurement of alcohol in blood using alcohol-test or other devices. This research aims to determine the effect of external agents such as alcohol on the CNS and how this effect is showed in behaviour changes on pupils and irises diameters and movements.

This paper proposes a Fused Capsule Network-based algorithm to automatically detect alcohol consumption from data extracted from periocular NIR images. The main contributions of this work are the following:

\begin{enumerate}
\item A novel dataset of 3,000 NIR periocular images from 30 volunteers was captured, which contains sessions with and without the influence of alcohol. This dataset will be available to other researchers upon request \footnote{\url{https://github.com/jedota/Iris_alcohol_classification}}.
\item A novel Capsule Network-based algorithm that detects the influence of alcohol on Periocular NIR images is proposed. The algorithm is called Fused-Capsule Network (F-CapsNet). It includes a convolutional block that extracts features from each data class separately and then fuses them into a two-layer capsule net. The resulting algorithm improves the performance of the traditional Capsule Network while using fewer parameters.
\item Additional classifiers such as SVM, Small-VGG, and feature extraction using a pre-trained model (embedding) were also implemented and tested for comparison and baseline. 
\end{enumerate}

The remainder of this paper is structured as follows: The related work is reviewed in Section \ref{SOA}. The Fusion Capsule Network algorithm and the database captured are described in Section \ref{FCN}. Experiments and conclusion are reported in sections \ref{Experiments} and \ref{Conclusions} respectively. 
\vspace{-0.2cm}

\section{Related Work}
\label{SOA}
\begin{figure}[]
\begin{centering}
\includegraphics[scale=0.35]{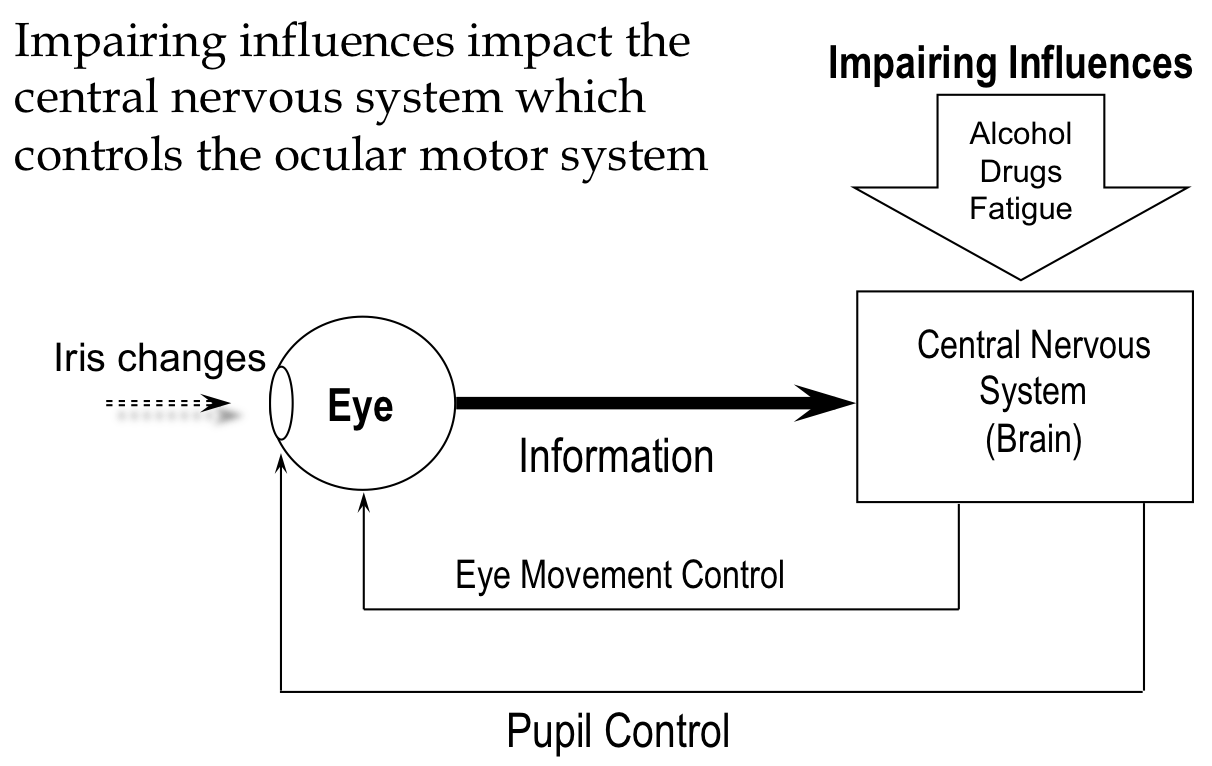}
\par\end{centering}
\caption{\label{diagrama} Block diagram of the influence of the alcohol in the Central Nervous System.}
\end{figure}

This section reports a brief review of alcohol influence in iris changes and its effects on "Fitness for Duties." Alcohol (ethyl alcohol or ethanol) is mainly used for recreational purposes and is the second most widely used drug after caffeine. It is essential to point out that the FFD does not relate to the traditional alcohol test. Here, we are detecting the effect on CNS and alertness of the worker due to alcohol consumption and not the Blood Alcohol Concentration (BAC). (See Figure \ref{diagrama}).
Alcohol consumption affects pupil changes directly, such as dilation. This change can be detected using state-of-the-art computer vision techniques, including Deep Learning. 

Alcohol is depressant for the Central Nervous System (CNS). It diminishes environmental awareness, response to sensory stimulation, cognitive functioning, spontaneity, and physical activity. See Figure \ref{diagrama}. Alcohol can produce increasing drowsiness, lethargy, amnesia, anti-epileptic effects, hypnosis, and anaesthesia in high doses. It is, therefore, not surprising that most countries restrict people from driving and operating dangerous equipment under the influence of alcohol (when blood alcohol concentration exceeds a certain level. For example, 0.05\% in Australia and 0.08\% in the United Kingdom \footnote{\url{http://www.who.int/gho/alcohol/en/}}).

Alcohol abuse directly affects workplaces with increased worker absenteeism, reduced productivity, increased working tardiness, frequent stoppages, lower quality work, increased number of accidents causing injury, and equipment damage\cite{wickwire2017shift, Frone, DorrianandSkinner, leo_causa}.

Navarro et al. \cite{Navarro7877181} developed a system that captures the driver iris image to detect if the person is drunk. That paper comprises a hardware and software system that implements an algorithm based on the Gabor Filter. The system consists of a Charge-Coupled Device (CCD) Camera and Analog-to-Digital Converter linked into a program to simulate the captured image. The system provides a signal to interact with the car ignition if the software detects that the driver is under alcohol.

Monteiro et al. \cite{MonteiroPinheiro2015} proposed a non-invasive and simple test to detect the use of alcohol through pupillary reflex analysis. Results present detection close to 85\% of accuracy using algorithms for pattern recognition. These results demonstrate the efficacy of the test method. The main limitation of this work is related to the active participation of the volunteers; each volunteer had to stay in a dark testing room for approximately 5 minutes to adapt the pupil dilation/constriction to the darkness.

Arora et al. \cite{AroraVatsaSinghEtAl2012} presented a preliminary study about the impact of alcohol on an Iris recognition system. The experiments were performed on the 'Iris Under Alcohol Influence' database. Results show that when comparing pre and post-alcohol consumption images, the overlap between mated and non-mated comparison score distributions increases by approximately 20\%. These results were obtained using a relatively small database (220 pre-alcohol and 220 post-alcohol images obtained from 55 subjects). The subjects consumed about 200 ml of alcohol (with 42\% concentration level) in approximately 15 minutes and the images were captured 15-20 minutes after alcohol consumption. That work suggests that about one in five subjects under the influence of alcohol may fail identification by iris recognition systems.  

Bernstein et al. \cite{BernsteinMendezSunEtAl2017} used spectrogram images of size $224\times224$ from audio waveforms to identify the presence of alcohol with Convolutional Neural Networks (CNN) and wearable sensors. They used 80 training images (40 positives, 40 negatives) and 20 test images (10 positives, 10 negatives) spectrograms, obtained promising results for non-audio waveforms.

Koukio et al. \cite{Koukiou,facialTermal} proposed the use of thermal images to identify individuals under the influence of alcohol. They have shown that changes in the eye temperature distribution in intoxicated individuals can be detected using thermal imagery \cite{eyetermal}. 

\section{Proposed Method}
\label{FCN}
This work proposes a modified Capsule Network architecture \cite{sabour} called Fusion CapNets (F-CapNets). It uses periocular NIR images to detect alcohol consumption. The architecture details are described in section \ref{fcapnet}. As an additional contribution, a novel periocular NIR database from volunteers with and without the influence of alcohol was captured. The images were analysed in order to show the influence of alcohol on the deformation of the pupil (Section \ref{databasedetails}). 

\subsection{Fusion Capsule Network (F-CapNets)}
\label{fcapnet}

Deep Learning techniques through Convolutional Neural Networks have been demonstrated to be a valid alternative to replace handcrafted methods for feature selection. CNN algorithms can learn specific features automatically. However, they usually require large volumes of data for training the network. Furthermore, the quantity of data required increases as the architecture goes deeper. 

\begin{figure}[]
\begin{centering}
\includegraphics[scale=0.3]{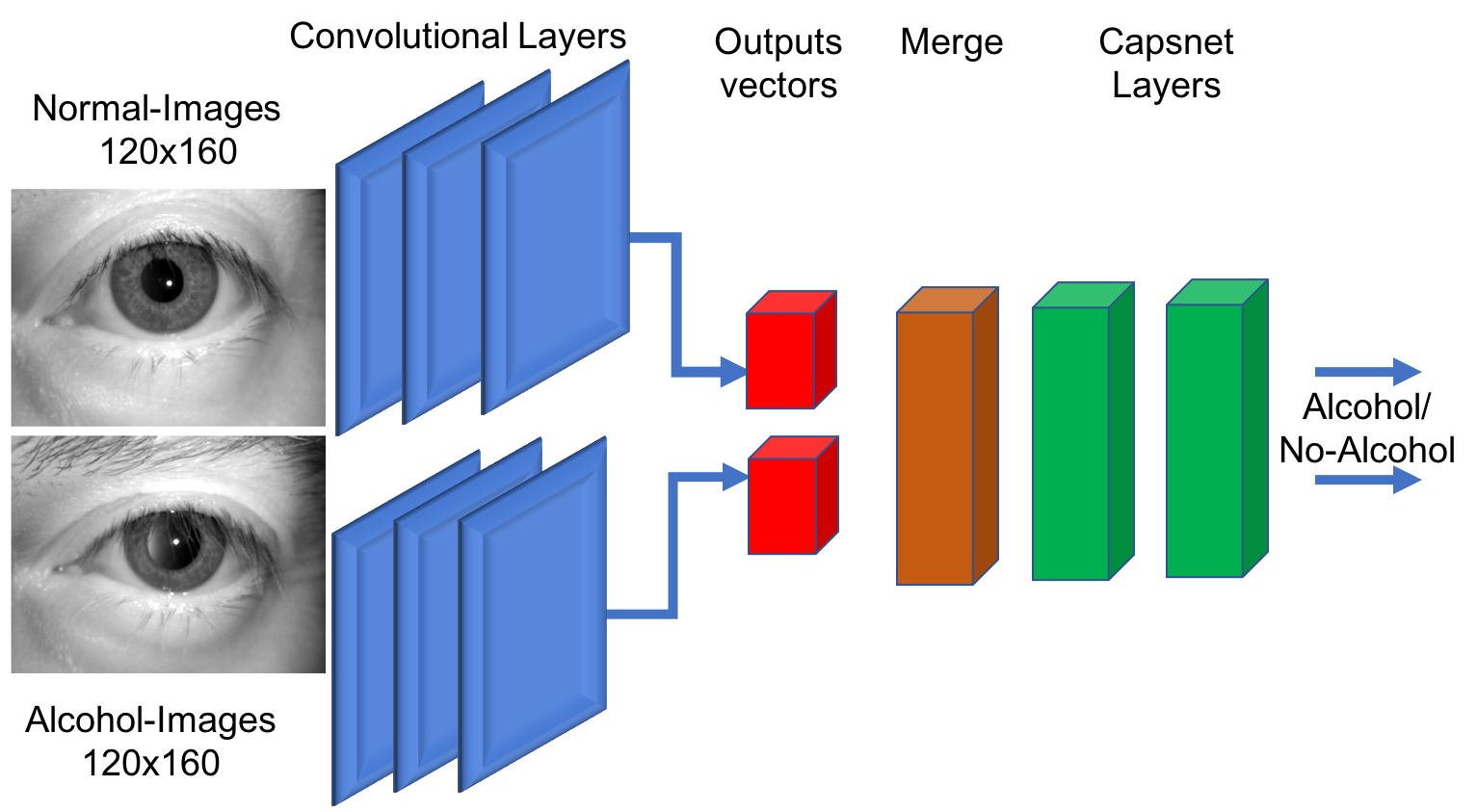}
\caption{ \label{archcap} Representation of train CapNets architecture proposed for alcohol classification. The convolutional layers extract information separately for alcohol and no-alcohol images and concatenate them before enter to the regular CapNets. For inference time, only one periocular image (potentially under alcohol) is sent to the system.}
\par\end{centering}
\end{figure}

In order to overcome the limitations of CNNs to work with reduced volumes of data and difficulties of handling changes in rotation or translation of the input data, a Capsule Neural Network is proposed. Despite traditional implementation, the proposed Fusion Capsule Network algorithm includes a pre-step where the feature extraction process is done separately for a pre-trained network for each class (Alcohol and No-alcohol images). A convolutional network block comprises the convolutional layers and one fusion layer. The extracted features for each category are then concatenated and used as input for the capsule network. The capsule block includes two layers with $N$ capsules. A scheme of the architecture used in the training process is shown in Figure \ref{archcap}. Only one periocular image (potentially under alcohol) within the two eyes on it is sent to the system for inference time.

This architecture helps to reduce the number of network parameters, making the algorithm suitable for implementation in mobile devices. In order to train and test the algorithm, a novel database was captured, and it is described as follows.

\subsection{IAL-I Database}

\label{databasedetails}
To the best of our knowledge, there is no publicly available database of NIR periocular images captured from volunteers under the influence of alcohol.
A novel database (IAL-I) from 30 volunteer subjects, 24 male and six female from the 25-50 age group, was captured in this work. The capture protocol and the images' analysis are presented as follows. A health committee team evaluated the captured process before it started. As requested by the health committee, a consent form was used to capture images for all the volunteers. This database will be available on request.
 
\subsubsection{Data Capture}
The capture process cooperates and delivers periocular images with both eyes. On average, 20 frames are captured per subject. This capture process takes five seconds. 
It is essential to highlight that the subject in the presence of alcohol shows an involuntary movement on X-axis because they cannot keep the position. This movement adds blurring to the captured images. The author reported the problems associated with eye detection and segmentation under alcohol effects in a different publication \cite{tapia2021semantic}. 

A TD-100 and Iritech NIR capture devices under a controlled environment were used. Marks on the floor at 30 cm up to 50 cm distance from the camera were used to facilitate the data capture process. 

Each subject was requested to step on the first mark (50 cm from the camera) and to look at the NIR sensor. An image of both irises was captured. A second image was taken with the subject placed at the second mark (30 cm from the camera). This process was repeated five times for each volunteer.

On the fifth time, a sequence of 20 consecutive frames from both eyes was captured to record changes in the pupil due to the light used by the NIR sensor.
This image sequence allows detection velocity changes of the iris across all the frames to be measured and, therefore, helps estimate the influence of alcohol on the volunteers. Alcohol directly affects the velocity of iris adjustment to direct light.  

The data capture process with both devices was organized into 5 sessions (According to our protocol):

\begin{itemize}
    \item Session 0: Images were captured when volunteers were not under the influence of alcohol.
    \item Session 1: Images were captured 15 minutes after the volunteers consumed 200 ml of alcohol with a concentration level of 42\%.
    \item Session 2: Images were captured 30 minutes after alcohol consumption.
    \item Session 3: Images were captured 45 minutes after alcohol consumption.
    \item Session 4: Images were captured 60 minutes after alcohol consumption.
\end{itemize}
The room temperature and lighting (200 lux) were kept constant in the capturing process. A total of ten images per eye were captured for each volunteer per session. The total number of images in each session is shown in Table \ref{database}.

\begin{table}[H]
\centering
\scriptsize
\caption{\label{database} Database description.}
\label{my-label}
\begin{tabular}{|c|c|c|c|l|c|}
\hline
\textbf{Session} & \textbf{Condition} & \textbf{Time} & \textbf{Alcohol} & \multicolumn{1}{c|}{\textbf{Images}} & \textbf{Total} \\ \hline
S0               & Pre-alcohol        & 0             & 0                & 20 (L-R)                   & 600              \\ \hline
S1               & Post-alcohol       & 15            & 200 ml           & 20 (L-R)                   & 600             \\ \hline
S3               & Post-alcohol       & 30            & 0                & 20 (L-R)                   & 600             \\ \hline
S4               & Post-alcohol       & 45            & 0                & 20 (L-R)                   & 600             \\ \hline
S5               & Post-alcohol       & 60            & 0                & 20 (L-R)                   & 600             \\ \hline
\end{tabular}
\end{table}

\begin{figure}[h]
\begin{centering}
\begin{tabular}{cc}
\includegraphics [scale=0.15]{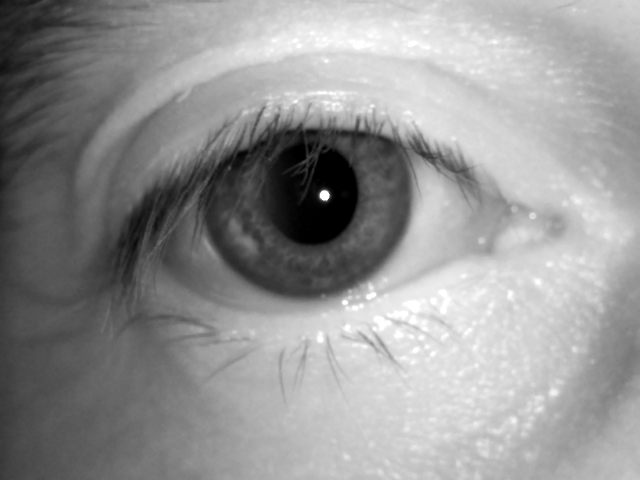}  \includegraphics [scale=0.15]{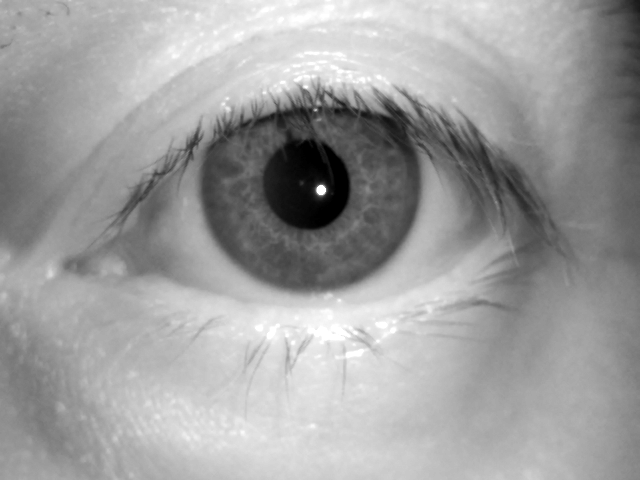}\tabularnewline
\includegraphics [scale=0.15]{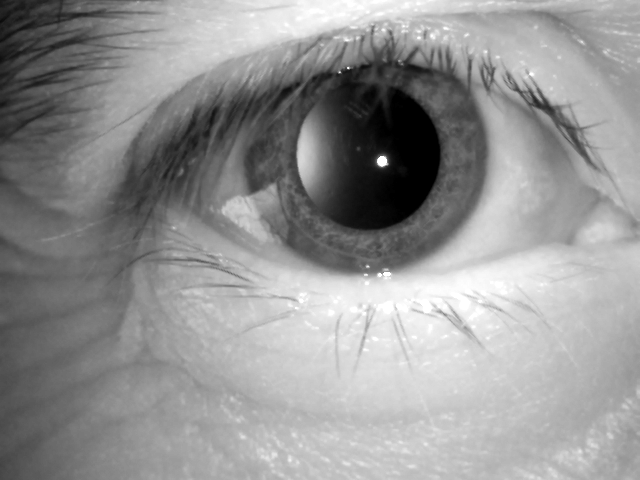} \includegraphics [scale=0.15]{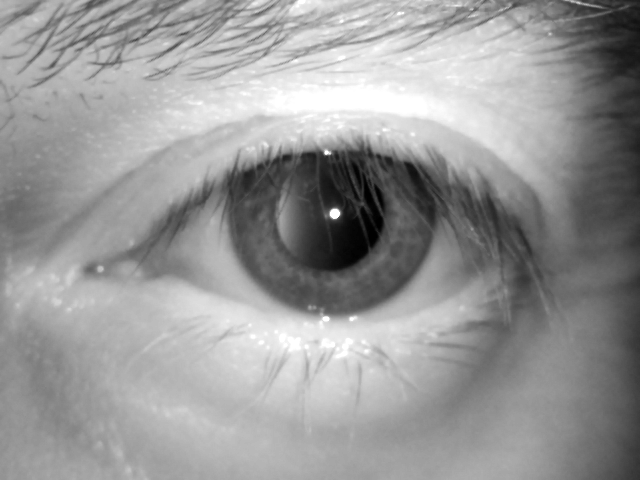}\tabularnewline

\end{tabular}
\par\end{centering}

\caption{\label{fig:figceleb} Top: Example of eyes in control condition. Below: Eyes under the effect of alcohol 30 minutes after consumption. The images belong to the same subject.}
\end{figure}

A total of 600 images of volunteers not under the influence of alcohol were captured and 2,400 images were taken after each volunteer had ingested 200 ml of alcohol (Images taken in intervals of 15 minutes after consumption). The database was divided into 70\% and 30\% for Training and Testing. The partition is a subject-disjoint database. Example images are shown in Figure \ref{fig:figceleb}.

\subsubsection{Image analysis}

According to the literature \cite{AroraVatsaSinghEtAl2012}, alcohol influence affects the size of the pupil and iris. It involves, in particular, the relation between both measures: iris and pupil radius. This relation is called $p$, and the values were normalized between values of 0.0 up to 1.0 according to the following expression:

\begin{equation}
 p(A)= Ir/Pr
\end{equation}

Where $A$ is the image, $p(A)$ represents the ratio between $Ir$ (iris radius) y $Pr$ (pupil radius). This measure can be used to compare the standard deviation between the radius with alcohol $p(A)$ and the radius without alcohol $p(NA)$. If the value $p(A)>X$ is considered as Dilation, and $p(A)<Y$ represents Contraction. With $X$ the average ratio of volunteers under control conditions is expressed (without any alcohol consumption).

A change can be observed in the dataset captured 30 minutes after the subjects consumed alcohol. However, the ratio of pupil size did not change drastically in all other sessions. The ratio of the pupils alone can not be considered as an indicator of the presence of alcohol. In fact, the distributions are overlapping. It is impossible to easily separate both distributions using only a threshold or a traditional machine learning algorithm.  

\begin{figure*}[]
\centering
\begin{tabular}{c}
\includegraphics [scale=0.42]{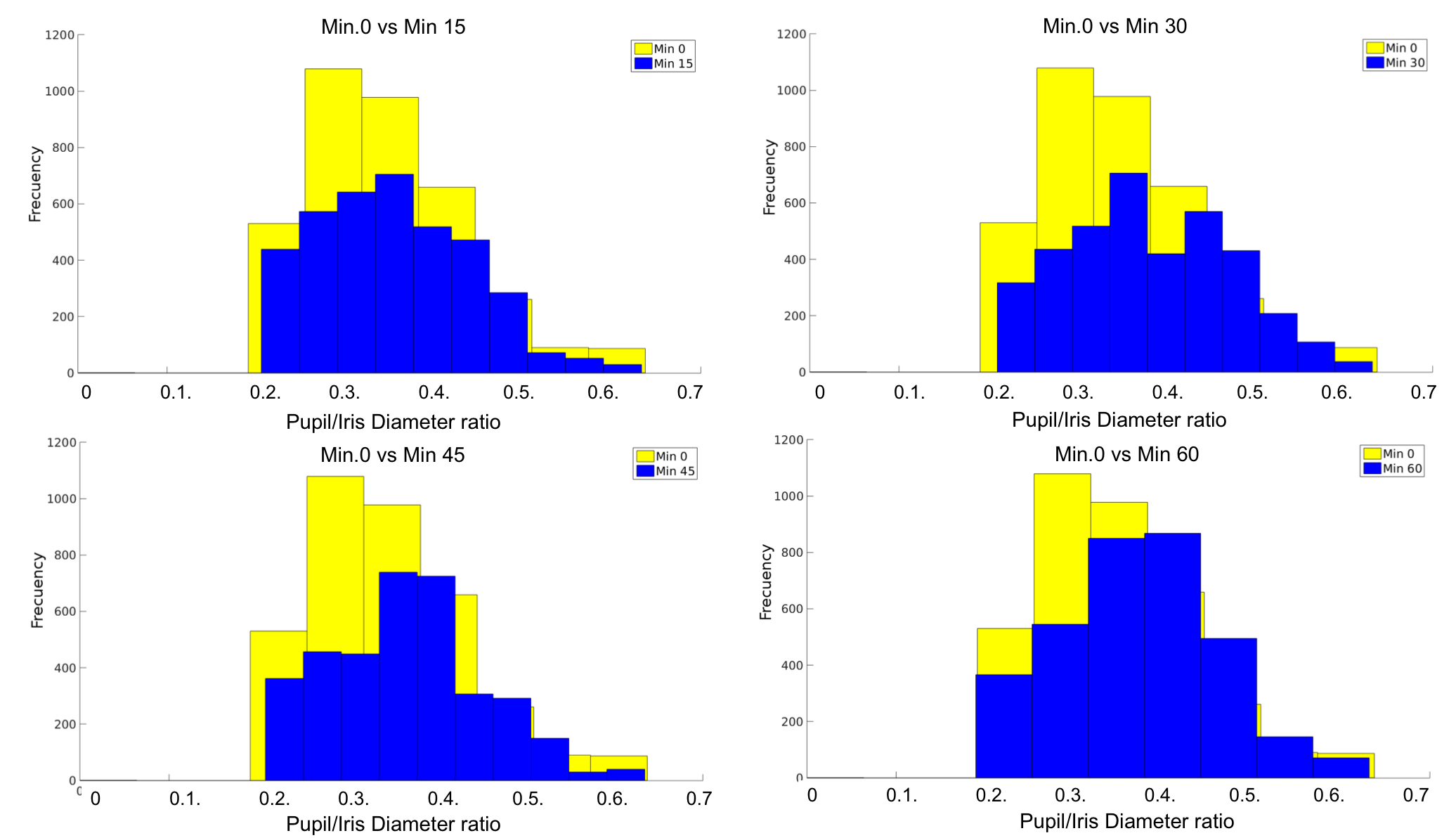} 
\end{tabular}
\caption{\label{fig:fig_barras} Pupils versus Irides ratio histograms for each alcohol captured sessions separately from 0, 15, 30, 45, and 60 minutes (Blue). Yellow represents Minute 0.}
\end{figure*}

Figure \ref{fig:fig_barras} Shows the histograms of $p(A)$ computed for images captured in the different sessions. The pupil and iris radius were computed using a semantic segmentation proposed in \cite{tapia2021semantic} to measure the $p$ ratio automatically. 

\section{Experiments and Results}
\label{Experiments}

The captured database (IAL-I) was used to train and test all the experiments. It was divided into two classes: Alcohol and No-alcohol. 

All the eyes images are cropped as shown in Figure \ref{fig:figceleb}. Each periocular NIR image (left and right) has a resolution of $120\times160$ pixels. Each image was converted into a vector and included in a matrix as $M$ $(1 \times 19,200)$ for the SVM and Small-VGG network. Each row represents one image, and each column in $M$ represents one feature. Also, the ArcFace \cite{deng2018arcface} pre-trained network was used for extracting embedded features (512-D) for F-CapsNet as an input of capsnet. 

Experiments with and without data augmentation were performed. This technique helps to increase the number of images available for training, obtaining a model with better generalization properties. An image generator based on imgaug algorithm \cite{imgaug} is used to create samples from the left and the right eyes, preserving the partition of train, test, and validation sets were used. When using data augmentation, the dataset was increased from 3,000 images to 12,000 images for each eye. The following geometric transformations used a rotation range of 10 degrees, width shift range and height range of 0.2, and a Zoom Range of 15\%. All changes were made using a 'Nearest' fill mode, meaning the images are taken from the corners to apply the transformation. The mirroring operation was not applied because this can 'transform' the left eye into a right eye considering the pixel's positions. 

Four sets of experiments are reported in section \ref{Experiments}. First, a baseline using periocular images as input with SVM and Small-VGG were tested. The second uses features extracted from the embedding vector from a pre-trained ArcFace model. This embedding vector was used as an input of SVM. The third used embeddings vector from ResNet50/100 \cite{resnet} and MobileNetV2 \cite{mobilenetv2}. The fourth uses periocular input images as input in a traditional capsnet and the proposed fusion capsule network (F-CapsNet) uses features extracted from the ArcFace as an input of the capnets.



\subsection{Images classification using SVM and CNN}

In order to compare the performance of the proposed Fusion Capsule Network algorithm, two additional classifiers (SVM and CNN) were implemented and applied to the same database.

\textbf{SVM Classifier:} A SVM classifier with Gaussian kernel was trained using LIBSVM implementation \cite{Chang:2011}. A 5-fold cross-validation was used on 60\% of the original data to select the best parameters. 

The selected model was trained on the full 60\% training data. Finally, the model evaluation was performed on the test data (40$\%$ of the dataset). The best result only reached 63.55\% +/- 0.9 of accuracy in classifying alcohol and no-alcohol influence from NIR periocular images.
\vspace{+0.3cm}

\textbf{Small VGG - CNN Classifier:}

Small-VGG network with three convolutional blocks and two fully connected layers with a small number of neurons was used. The choice of a smaller network design was motivated by the desire to reduce the risk of overfitting.  The network processed only one channel (grayscale image of size $120\times160$). Sparse, convolutional layers and max-pooling are at the heart of the Small-VGG models. The three convolutional blocks are defined with 128, 256 and 512 filters, respectively.

Several experiments for the fusion of left and right periocular images were performed according to the parameters reported in Table \ref{leftEye}. In the table, the classification accuracy of the algorithms is also reported. The best result obtained was 73.41\%.
\vspace{-0.3cm}

\begin{table}[H]
\centering
\scriptsize
\caption{Periocular Results on Small-VGG. Drop-out$=0.5$, Acc. Represents the Accuracy. BS represents Batch size.}
\label{leftEye}
\begin{tabular}{|c|c|c|c|c|c|c|}
\hline
\textbf{Epoch} & \textbf{BS} & \textbf{C1} & \textbf{C2} & \textbf{C3} & \textbf{DENSE} & \textbf{Acc. (\%)} \\ \hline
100  & 16          & 32       & 32   & 32  &             7200        & \textbf{73.41}      \\ \hline 
100  & 16          & 16       & 16   & 32  &             7200        & 70.07           \\ \hline 
100  & 16          & 8        & 8    & 16  &             7200        & 59.02           \\ \hline  
100  & 16          & 4        & 4    & 8   &             7200        & 65.90           \\ \hline 
100  & 16          & 32       & 16  & 16   &             7200        & 68.30           \\ \hline
\end{tabular}
\end{table}


\subsection{Embedding}
In previous sections, the images are used to input classifiers using a Small-VGG and SVM. However, in the state-of-art, it was shown that also embeddings could be used to train a classifier. An embedding vector of real numbers contains all extracted information from a pre-trained network. According to our revision, there are no available pre-trained networks under alcohol presence. Therefore, the ArcFace recognition model was used to extract the features. This feature vector has 512 elements. Table \ref{tab:emb}, shows the classification results using embedding information extracted from ArcFace. The best result was 73,0\% when the embedding was extracted from the 45 convolutional layers. This result is very similar to Small-VGG. Despite the fact that the embedding may help to extract information, such information is not enough to distinguish pre-alcohol and post-alcohol images. Fine-tuning was also explored without success and with lower power of generalization.
\vspace{-0.3cm}

\begin{table}[H]
\centering
\caption{Classification results from embedding features extracted from ArcFace.}
\label{tab:emb}
\begin{tabular}{|c|c|c|c|c|c|c|}
\hline
\textbf{Classifier} & \textbf{Layers} & \textbf{TP(\%)} & \textbf{TN(\%)} & \textbf{FP(\%)} & \textbf{FN(\%)}& \textbf{Acc.(\%)} \\ \hline
SVM             & 45              & 93.0    & 54.8    & 45.2   & 7.0&    \textbf{73.0} \\ \hline
SVM             & 50              & 86.2    & 48.7    & 51.3.  & 13.8&   67.0 \\ \hline
SVM             & 54              & 91.6    & 49.8    & 50.2   & 8.4&    70.0 \\ \hline
\end{tabular}
\end{table}

\subsection{Images classification using Pre-trained Networks}

Several models have been proposed in state-of-the-art used face images. However, deep face recognition networks have shown that they can work very robustly, even on challenging data. Therefore, we also propose to employ deep face representations extracted by such deep face recognition systems for alcohol classification \cite{deepfake}. As we mentioned before, it would be possible to apply transfer learning and re-train a pre-trained deep face recognition network to detect alcohol images. However, the high complexity of the model, represented by the large number of weights in the neural network, requires a large amount of training data. Even if only the lower layers are re-trained, the limited number of training images (and a much lower number of subjects) in our database can easily result in over-fitting to the characteristics of the training set.
Three pre-trained networks were used such as MobileNetv2 \cite{mobilenetv2} and ResNet50/100 \cite{resnet}. All models are based on the Imaginet database~\cite{imagenet} and were used to extract features on the lowest layer. The eye images were resized to $224\times224\times3$. The output sizes before flatten layer are 2,048 for ResNet and 49 for MobileNetv2, respectively. Table \ref{tab:deep} shows that ResNet50 obtained the best results with 73.8\%. This result is very similar to the ArcFace results on Table \ref{tab:emb}. All the results are presented in Table \ref{tab:deep}.

\begin{table}[H]
\centering
\caption{Classification results from three pre-trained networks.}
\label{tab:deep}
\begin{tabular}{|c|c|c|c|c|c|c|}
\hline
\textbf{Classifier}  & \textbf{TP~(\%)} & \textbf{TN~(\%)} & \textbf{FP~(\%)} & \textbf{FN~(\%)}& \textbf{Acc.(\%)} \\ \hline
MobilNetV2           & 91.0    & 60.0    & 47.0   & 10.0  & 72.6 \\ \hline
ResNet50             & 88.0    & 62.0    & 45.0.  & 8.0.  & 73.8 \\ \hline
ResNet101            & 86.0    & 58.0    & 46.0   & 9.0  & 72.3 \\ \hline
\end{tabular}
\end{table}

\subsection{Images classification using Capsule Network}

In this section the traditional capsule network and the proposed F-CapsNet algorithms are tested and compared.
\newline
\newline
\textbf{Traditional Capsule Network algorithm:} The algorithm proposed by \cite{sabour} was implemented for this experiment. Two layers with several capsules number were tested (8, 16, 32, and 64). Experiments with and without data augmentation were also performed. A grid search was used to look for the best parameters for capsules and routing (8, 16, 32, and 64 capsules and 3, 4, and 5 routes). The output has the original image size $120 \times 160$ with sigmoid activation. 
The reconstruction error was optimized to increase model accuracy and avoid the model over-fitting to the training data. Thus, the reconstruction error was set to 0.0005. A grid search involving six options for the reconstruction error (from 0 to 0.00005, dividing by 10 in each step), selecting from 64, 128, or 256 features maps as the output of each convolution and picking among five options for the kernel size of the convolutional layers (3, 5, 7, 9 and 11).  Learning rate values were explored from $0.1$ up to $1e-5$. The best value was reached at $1e-5$.

\textbf{Proposed Fusion Capsule Network algorithm (F-CapsNet):}
The proposed architecture (As shown in Figure \ref{archcap}) uses a pre-trained convolutional block to extract the features separately from each class (Alcohol and no-alcohol). The output of the convolutional layers with the best features for alcohol and no-alcohol images are merged before going through the CapNets layer. This process allows the number of parameters of the model to be reduced (See Table~\ref{tab:res_cap}).

F-CapsNet uses a decoder (at the end of the last capsule) to regularise and reconstruct the original iris images. 
The first layer with 8 capsule has $9,600$ vectors ($80\times60\times2$), ReLU activation, $256$ filters, kernel size of $3 \times 3$ and,  stride size of  $1$.  The second layers with 8 capsule has 1,024 neurons, ReLU activation, 512 filters, strides $2$, kernel size of $3 \times 3$. 

Four different options were considered for the number of primary capsules and their dimensions (8, 16, 32, and 64) according to the number of feature maps. The number of routing iterations was tuned by considering $5$ different options (sequence from $1$ to $5$ with a step of $1$). The best results obtained are reported in Table \ref{tab:res_cap}. The proposed model Caps8-Fusion reaches a high accuracy ($92.35\%$) when detecting the influence of alcohol in periocular NIR images. The resulting model contains fewer parameters than the traditional capsule network implementation (reduction of approximately $50\%$ of the number of parameters). 



Table~\ref{tab:res_cap} shows the results obtained for each experiment. The first column indicates the name (CapsX) of the model. Where $X$ represents the number of capsules used for each model. The expression DA in the model's name indicates the use of data augmentation. The Caps8-Fusion reaches the best accuracy with a low number of parameters $(9M)$ in comparison of traditional CapsNet $(33M)$. Best results are highlighted in Bold.

\begin{table}[]
\centering
\scriptsize
\caption{Best results of CapsNet implementation. DA. represents Data-Augmentation. \# represent numbers.}

\label{tab:res_cap}
\begin{tabular}{|c|c|c|c|c|c|}
\hline
\textbf{Model}                                                     & \textbf{\begin{tabular}[c]{@{}c@{}}\#\\ Caps\end{tabular}} & \textbf{\begin{tabular}[c]{@{}c@{}}Accuracy\\ (\%)\end{tabular}} & \textbf{Parameters} & \textbf{\begin{tabular}[c]{@{}c@{}}Specificity\\ (\%)-TNR\end{tabular}} & \textbf{\begin{tabular}[c]{@{}c@{}}Sensitivity\\ (\%)-TPR\end{tabular}} \\ \hline
Caps8                                                              & 8                                                             & 90.08                                                           & 18,066,220          & 87.06                                                                  & 93.10                                                                  \\ \hline
Caps16                                                             & 16                                                            & 90.15                                                           & 20,249,800          & 88.21                                                                  & 92.09                                                                  \\ \hline
Caps32                                                             & 32                                                            & 90.26                                                           & 24,615,744          & 89.15                                                                  & 91.37                                                                  \\ \hline
Caps64                                                             & 64                                                            & 91,35                                                           & 33,348,416          & 89.95                                                                  & 92.76                                                                  \\ \hline
\begin{tabular}[c]{@{}c@{}}Caps8\\ DA\end{tabular}                 & 8                                                             & 91.15                                                           & 18,066,220          & 90.05                                                                  & 93.16                                                                  \\ \hline
\begin{tabular}[c]{@{}c@{}}Caps16\\ DA\end{tabular}                & 16                                                            & 90.26                                                           & 20,249,800          & 90.15                                                                  & 93.05                                                                  \\ \hline
\begin{tabular}[c]{@{}c@{}}Caps32\\ DA\end{tabular}                & 32                                                            & 88.15                                                           & 24,615,744          & 87.25                                                                  & 89.05                                                                  \\ \hline
\begin{tabular}[c]{@{}c@{}}Caps64\\ DA\end{tabular}                & 64                                                            & 86.25                                                           & 33,348,416          & 86.35                                                                  & 86.15                                                                  \\ \hline
\textbf{\begin{tabular}[c]{@{}c@{}}Cap8\\ Fusion\end{tabular}}     & \textbf{8}                                                             & \textbf{92.35}                                                           & \textbf{9,060,220}  & \textbf{91.91}                                                                  & \textbf{92.79}                                                                  \\ \hline
\begin{tabular}[c]{@{}c@{}}Caps8-DA\\ Fusion\end{tabular} & 8                                                             & 92.26                                                           & 9,060,220  & 91.33                                                                  & 93.19                                                                  \\ \hline
\end{tabular}
\end{table}

Table \ref{tab:res_cap} also reports the True Positive Rate and the Negative Rate, which are standard metrics for binary classification. For this paper, the TNR refers to the accuracy of the negative class (No-Alcohol), and the TPR is the accuracy of the positive class (Alcohol). The alcohol prediction class shows to be more confident since it has a high TPR metric value and a low TNR level. 

\begin{figure}[]
\begin{centering}
\begin{tabular}{cc}
(a)\includegraphics [scale=0.37]{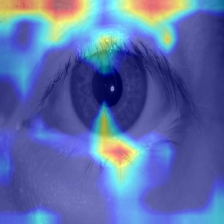} 
(b)\includegraphics [scale=0.37]{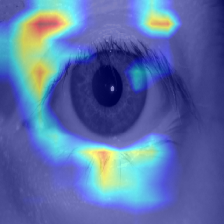}\tabularnewline
(c)\includegraphics [scale=0.37]{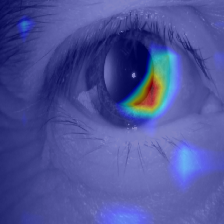} 
(d)\includegraphics [scale=0.37]{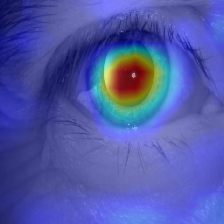}\tabularnewline
\end{tabular}
\par\end{centering}
\caption{\label{fig:figgradcam} Figures (a) and (b) show examples of eyes under control conditions. Below: (c) and (d) eyes under the influence of alcohol. Figures (a) and (c) belong to CNN outputs. And (b) and (d) belongs to F- Capsule network outputs.}
\end{figure}

Figure \ref{fig:figgradcam} shows four heatmaps of the most relevant areas that the CNN and F-capsule network algorithms are using for classification. The Gradcam ++ visualisation algorithm \cite{gradcam} was used. Red and blue colors represent higher and lower responses to the model, respectively. Areas with lower responses are less relevant to classification. The higher activation area in images taken from subjects under the influence of alcohol is located mainly in the iris. Conversely, the activation area for images from volunteers without alcohol consumption is located in the periocular region of the iris. People under alcohol effects tend to close the eyes; then changes will appear in the skin periocular areas. 
The relevant areas are represented for an average image and show the average region highlighted from 15 to 60 minutes images. The site over the iris captures the changes of the pupil and texture in different dilation positions for the same person. This prediction is possible because of the multiple iris images for each subject.

\section{Discussion and Conclusions}
\label{Conclusions}

As a relevant conclusion, we can point out that it is possible to detect alcohol consumption using NIR iris images. Also, this work shows that the ratio of the pupils alone for one image can not be considered as a sufficient indicator of the presence of alcohol. The distributions are overlapping, as shown in Figure \ref{fig:fig_barras}. Thus, it is impossible to separate both distributions easily using only a threshold or a traditional machine learning algorithm.
The correct alcohol presence is feasible to detect because we capture the pupil and iris texture variation across 60 minutes. 

SVM and CNN classifiers have been shown to obtain low results
when attempting to classify alcohol consumption from NIR periocular images (63.5$\%$ and 73.4$\%$ of accuracy, respectively). In the CNN algorithm, there is a loss of information in the pooling layer. There is a positional invariance of the components that the network uses to carry out the classification, which prevents it from being sensitive to changes in rotation or translation within the image. 

A Fused Capsule Network algorithm was proposed (F-CapsNet). This algorithm achieved a classification rate of 92.3$\%$. The properties of each object in the image are expressed as a vector that is mapped and routed to the final image. It includes a CNN block that extracted features separately from each class and fused them to feed the capsule layers. 

Only half of the parameters (9M) of the standard CapsNet implementation (Caps8) instead of $18M$ parameters are needed to model F-CapsNet, making it suitable for implementation in mobile devices. See Table \ref{tab:res_cap}.

This work is part of ongoing research that aims to develop a robust algorithm to estimate 'Fitness for Duty' from periocular NIR images. This research aims to develop technologies to pro-actively detect volunteers under the influence of alcohol, drugs, and fatigue.

\section{ACKNOWLEDGMENTS}
This research work has been partially funded by the German Federal Ministry of Education and Research and the Hessian Ministry of Higher Education, Research, Science and the Arts within their joint support of the National Research Center for Applied Cybersecurity ATHENE and Universidad de Chile-DIMEC.

{\small
\bibliographystyle{ieeetr}
\bibliography{ICPR.bib}
}

\end{document}